%% file: arxiv_main.tex
\documentclass{article}

\usepackage{graphicx}
\usepackage{subcaption}
\usepackage{rotating}
\usepackage{changepage}
\usepackage{lmodern}

\usepackage[
    backend=biber,
    style=authoryear,
    sorting=none,
    doi=false,isbn=false,url=false,eprint=false,
    natbib=true,
    mincitenames=1,
    maxcitenames=2
]{biblatex}
\AtEveryBibitem{
 \clearlist{address}
 \clearfield{date}
 \clearfield{eprint}
 \clearfield{isbn}
 \clearfield{issn}
 \clearfield{month}
 \clearlist{organization}
 \clearfield{series}
 
 \ifentrytype{book}{}{
  \clearlist{publisher}
  \clearname{editor}
  \clearlist{location}
 }
}

\usepackage{xurl}
\usepackage{hyperref}
\hypersetup{
    hidelinks=true,
    breaklinks=true
}
\usepackage{textcomp}

\usepackage{booktabs}
\usepackage{nicefrac}
\usepackage{microtype}
\usepackage{xcolor}
\usepackage{numprint}

\usepackage{amssymb}
\usepackage{amsmath}
\usepackage{amsfonts}

\usepackage{algorithm}
\usepackage{algpseudocode}
\usepackage{threeparttable}

\usepackage{authblk}

\usepackage{geometry}
\input{notation}

\usepackage{ntheorem}
\theoremstyle{break}
\newtheorem{definition}{Definition}[section]

\newcommand{\frameindraft}[1]{\frame{#1}}
\renewcommand{\frameindraft}[1]{#1}

\begin{document}

\title{Collaborative Learning From Distributed Data With Differentially Private Synthetic Twin Data}

\date{}

\author[1]{Lukas Prediger}
\author[2,3]{Joonas Jälkö}
\author[2]{Antti Honkela}
\author[1,4]{Samuel Kaski}
\affil[1]{\scriptsize Department of Computer Science, Aalto University}
\affil[2]{\scriptsize Department of Computer Science, University of Helsinki}
\affil[3]{\scriptsize Work done while at Aalto University}
\affil[4]{\scriptsize Department of Computer Science, University of Manchester}

\maketitle

\begin{abstract}

    Consider a setting where multiple parties holding sensitive data aim to collaboratively learn population level statistics, but pooling the sensitive data sets is not possible. We propose a framework in which each party shares a differentially private synthetic twin of their data. We study the feasibility of combining such synthetic twin data sets for collaborative learning on real-world health data from the UK Biobank.
    We discover that parties engaging in the collaborative learning via shared synthetic data obtain more accurate estimates of target statistics compared to using only their local data. This finding extends to the difficult case of small heterogeneous data sets. 
    Furthermore, the more parties participate, the larger and more consistent the improvements become. Finally, we find that data sharing can especially help parties whose data contain underrepresented groups to perform better-adjusted analysis for said groups.
    Based on our results we conclude that sharing of synthetic twins is a viable method for enabling learning from sensitive data without violating privacy constraints even if individual data sets are small or do not represent the overall population well.
    The setting of distributed sensitive data is often a bottleneck in biomedical research, which our study shows can be alleviated with privacy-preserving collaborative learning methods.

\end{abstract}

\section{Introduction}
\label{sec:introduction}

Often access to the data needed for the most crucial statistical inference tasks is strictly limited to protect the privacy of data subjects. One example where this is especially prevalent is the case of medical data. Due to this limitation, such data cannot be easily combined across different origins to make population level statistical discoveries. This can be a severe problem during newly developing situations such as epidemics, in which data at each single origin is initially scarce and such discoveries are essential in making informed population-level decisions, for example regarding the measures to take to prevent infectious diseases from spreading. Recent advances in generative models as well as privacy-preserving machine learning make sharing synthetic data an appealing solution to mitigate the privacy concerns of sharing sensitive data.

Releasing synthetic data generated from a model trained with differential privacy \citep{DworkMNS06} - so-called \emph{synthetic twin data} - has been proposed previously as a way to enable privacy-preserving sharing of sensitive data sets \citep{HardtLM12,ChenAC12,Zhang+14,AcsMCD18,Abay+18,YoonJvdS18,McKennaSM19,Beaulieu+19,HarderAP21,JalkoLHTHK21,RaisaJKH22}.
These previous works have focused on methods for releasing a synthetic twin data set for a single large sensitive data set. However, the usefulness of combining synthetic twin data sets released by multiple parties for collaborative analysis has not been studied before. This is particularly true for the case where data are not homogeneous amongst the parties and each party holds a relatively small sensitive data set. We bridge this gap in the present work by performing a case study on a real world data set.

\begin{figure*}[tbh]
    \centering
    \begin{subfigure}[t]{.49\textwidth}
        \centering
        \includegraphics[width=.88\linewidth]{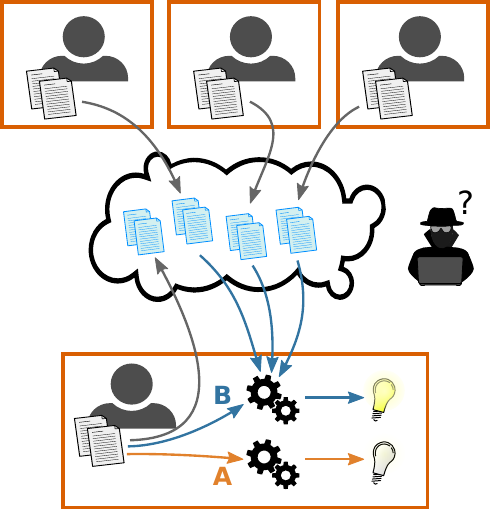}
        \label{fig:overview_illustration}
    \end{subfigure}
    \hfill
    \begin{subfigure}[t]{.49\textwidth}
        \centering
        \frameindraft{\includegraphics[width=\columnwidth]{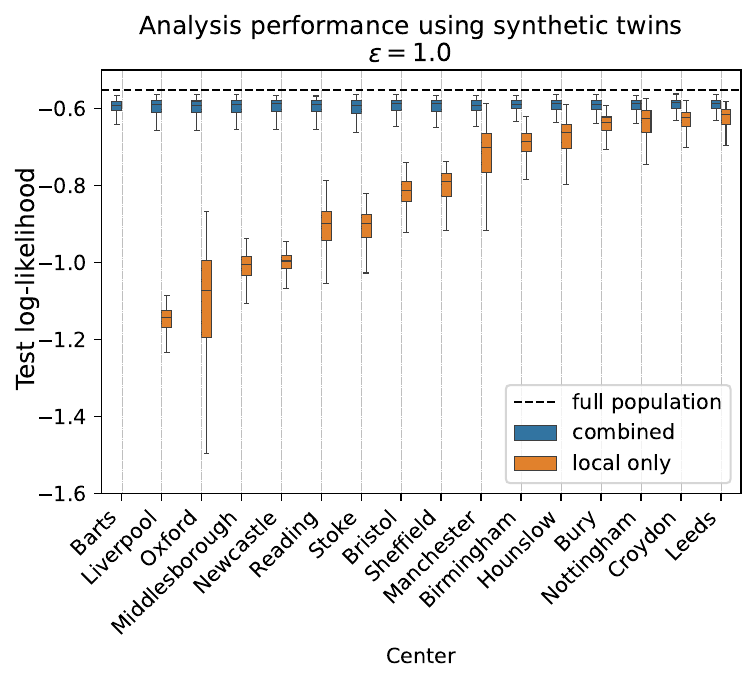}}
        \label{fig:log_likelihood_per_center}
    \end{subfigure}
    
    \caption{\textbf{Left:} Schematic overview of our setup. Multiple parties create synthetic twin data sets of their local data under privacy guarantees and make them publicly available. Any single party can then use the published synthetic data when performing a data analysis task (case B, blue) to improve results over only using their local data (case A, orange). The original data never crosses the (orange) privacy barriers. \textbf{Right:} Predictive log-likelihoods of the learned model (blue) are significantly improved over using only locally available data (orange). Uncertainty is also reduced significantly. The dashed black line shows the log-likelihood for an impractical ideal setting where the analysis could be performed over the combined data of all parties. Log-likelihood is evaluated on a held-out test set of the whole population and normalised by dividing with the size of the test set. The box plots show the distributions of log-likelihood for parameters sampled from the distributions implied by maximum-likelihood solution and errors obtained from the analysis task and over 10 repeats of the experiment. Boxes extend from $25\%$ to $75\%$ quantiles of the obtained log-likelihood samples, with the median marked in the box. Whiskers extend to the furthest sample point within $1.5$ inter-quartile range. Higher mean log-likelihoods of combined over local only are statistically highly significant ($p < 0.001, n_{\text{local only}}=\numprint{1000}, n_{\text{combined}}=\numprint{100000}$) for all centers. Local data log-likelihood of outlier center \emph{Barts} is cut off for improved readability (median: $-3.65$). The full figure can be found in App.~\ref{app:figures}.}
    \label{fig:abstract_figure_and_opening_result}
\end{figure*}

Throughout this work we consider a setting in which there exist $M \geq 2$ parties that are interested in performing statistical analyses over a population. Each party $m$ has access to a local data set $D_m$ that are \emph{disjoint} and \emph{non-uniformly} sampled from the overall population: Every $D_m$ may follow a distribution shifted away from that of the overall population (i.e., $\Pr[x | m] \neq \Pr[x]$). We assume that the parties cannot simply pool their data to perform the analysis due to the sensitive nature of the data. Instead, we suggest that each party trains a generative model on their local data using privacy-preserving machine learning techniques and publishes synthetic data sampled from this model in place of the sensitive data. The party then obtains synthetic data from other parties, which it combines with its own local data before performing its analysis task. This process is depicted in Fig.~\ref{fig:abstract_figure_and_opening_result}. More concretely, for training the generative model we adopt the formal framework of differential privacy (DP), which guarantees that the obtained generative model would be essentially identical if any data subject would be removed from the party's local data set. Since the parties' local data sets are disjoint, these privacy guarantees hold independently for all synthetic twin data sets. As a result the framework we describe achieves DP in the billboard model \citep{HsuJHRW16}, i.e., the local model obtained by each party is differentially private with respect to the data from all other parties. The formal privacy definition and generative model are detailed in Section~\ref{sec:methods}.

In our setting we assume that
\begin{enumerate}
    \item the parties' local data sets are disjoint,
    \item the parties' combined data accurately represent the overall population but any particular party's local data may be arbitrarily skewed,
    \item each party's goal is to optimise their analysis on the population level (not only their local data),
    \item the results of a party's analysis are kept private,
    \item a party that receives synthetic data will also share a synthetic twin of their own data, and,
    \item parties are non-malicious, i.e., they do not actively try to negatively affect other parties' performance.
\end{enumerate}

This setting leads to an apparent dilemma: If the local data of a party $m$ is not sufficient to learn the analysis model for the global population well, this suggests that it might not be possible to learn a good generative model from it either, especially under privacy constraints. Hence, we should expect that in the case where most parties only have access to small data sets and would therefore be interested in obtaining additional data to improve their analysis, the synthetic twin data sets shared amongst the parties might not be of sufficient quality to actually help. Specifically, the question arises: Does incorporating (low-fidelity) synthetic data generated from small data sets of other parties improve results of the analysis performed by party $m$ over just using its own (small) local data set? We answer this question in the affirmative.

Concretely, in our setting and under the assumptions stated above, we empirically demonstrate on a real-world health data set that:

\begin{enumerate}
    \item Complementing local data with synthetic twins of similarly sized data sets consistently increases the utility in the analysis task, and this increase can be drastic.
    \item This effect is more pronounced the smaller the local data is.
    \item As the number of parties sharing data increases, the results from the analysis on combined data quickly approach those obtained from the overall population.
    \item Parties suffering from local skew benefit from sharing, even if their local data is comparatively larger than that of any other party. Such skew can, for example, arise from underrepresented minority groups in a party's local data.
\end{enumerate}

The remainder of the paper is organised as follows: We first present the results of our empirical study in Section~\ref{sec:results}, followed by a discussion of the results and related literature in Section~\ref{sec:discussion}. Details about data and methods follow in Section~\ref{sec:methods}.

\section{Results}
\label{sec:results}
We empirically demonstrate our claims using data from the UK Biobank \citep{Sudlow+15} to learn a Poisson regression model for the risk of an individual person to test positive for SARS-CoV-2 based on socio-economic factors as the analysis task of interest, following a study performed by \citet{Niedzwiedz+20}. The data set splits naturally into 16 \emph{assessment centers} by geographic location, which we use as the local data sets available to 16 parties.
For each party we split off 20\% of the data prior to running the experiments as a held-out test set. These we combined into a \emph{global test set}, representative of the distribution of the full (combined) cohort. We describe the data set, analysis task and the generative model used to sample synthetic twin data for centers in Section~\ref{sec:methods}.

The UK Biobank data, and the per-assessment-center splits, present us with a large cohort of data. However, in many practical scenarios the parties would have significantly less data available. In order to simulate such a scenario, we will first use only subsamples of $10\%$ of the training data size for each center. Later, we will further experiment with larger subsample sizes as well as with the full data.

To summarise the quality of the regression model fit we use the distribution of predictive log-likelihoods on the test set that are induced by the maximum likelihood and standard error estimates. A better model will result in a log-likelihood distribution that is more concentrated at higher values.

\subsection{Data sharing consistently improves results over using local data only}
\label{sec:it_works}

We first show that a party $m$ which incorporates synthetic data shared by all other parties improves the performance of its analysis.

The right side of Fig.~\ref{fig:abstract_figure_and_opening_result} shows the log-likelihood of the Poisson regression model on the global test set for each center fitting the model only on its locally available data (orange) compared to incorporating synthetic data shared by all other centers (blue). 
Across all centers (subsampled to $10\%$ of the original size\footnote{\numprint{100} - \numprint{500} data points per center after sampling.}) we observe a clear improvement in average predictive log-likelihood when pooling synthetic data, as well as a reduction in spread. That is, we consistently obtain models that perform better and exhibit significantly less standard error in parameters. The log-likelihood distributions we obtain from including synthetic data are close to an ideal, privacy-agnostic baseline where we could simply pool all centers' data before performing our regression (black dashed line); this situation is precluded in practice due to privacy constraints.

The results appear consistent across centers even though the statistical signal for the regression model present in the local data varies drastically across the centers. This suggests that even centers whose local data already allows fitting a model that performs quite well on the population level test set, such as e.g., \emph{Nottingham}, \emph{Croydon} and \emph{Leeds}, still benefit from participating. This is true despite some of the other participating centers contributing data which alone result in a very poor model for predictions on the global population (such as e.g. \emph{Barts} and \emph{Oxford}).

\subsection{Gains increase quickly with number of shared data sets}
\label{sec:more_sharing_better}

We have seen in the previous section that every center improves the performance of its analysis task by incorporating large amounts of synthetic data from different sources. Indeed, due to the large amount of synthetic data available, the effect of the local data on the outcome of the analysis appears to be quite limited in that setting. It is now natural to ask how soon these improvements manifest, i.e., how much synthetic data is required. We investigate this in the following experiment: We fix a center and add synthetic data from other centers one by one, then evaluate the log-likelihood at every step. We repeat this experiment 100 times for different sequences in which synthetic data from other centers is added. 

\begin{figure}[t!bh]
    \centering
    \frameindraft{\includegraphics[width=.7\columnwidth]{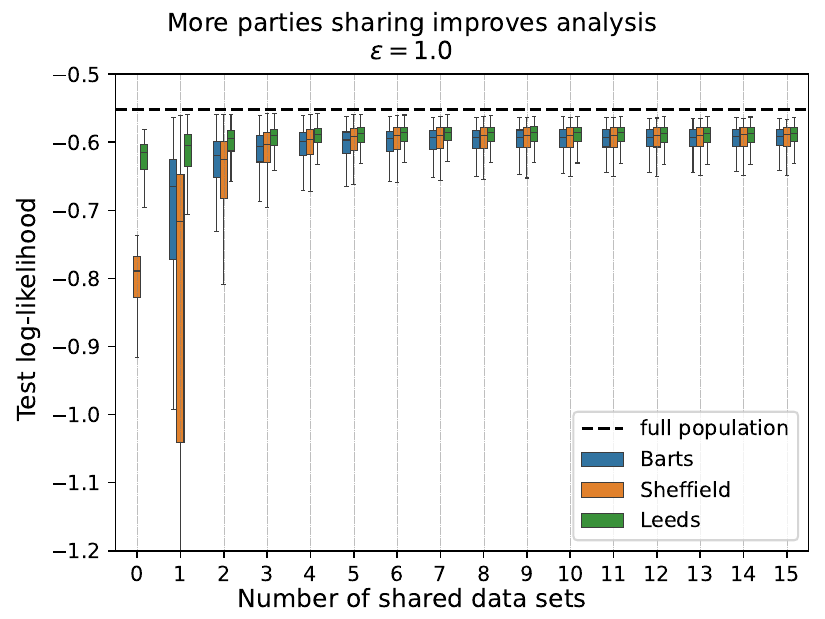}}
    \caption{The log-likelihood of the learned model improves rapidly as synthetic data from other centers becomes available. Spread in log-likelihoods may initially increase when only a few synthetic data sets are incorporated, but then diminishes rapidly with the number of additional data sets. The dashed black line shows the log-likelihood for an ideal setting where the analysis could be performed over the combined data of all parties. 10 repeats of the experiment, each with 100 repeats with different orders in which synthetic data is added. The improvements in mean log-likelihood between subsequent steps for all centers are highly significant ($p < 0.001, n=\numprint{100000}$) up to six centers releasing synthetic data. Local results for outlier center \emph{Barts} (median: $-3.65$) and whisker extension (to $-1.62$) of box for \emph{Sheffield} cut off for readability. See App.~\ref{app:figures} for corresponding p-values.} 
    \label{fig:ll_over_num_shared}
\end{figure}

Fig.~\ref{fig:ll_over_num_shared} shows box-plots of the log-likelihood distributions as more and more other centers make synthetic data available, from the perspective of centers \emph{Barts}, \emph{Sheffield} and \emph{Leeds}, representing respectively centers with bad, intermediate and good fit when using only local data (see Fig.~\ref{fig:abstract_figure_and_opening_result}, right). Results for the remaining centers are consistent with the ones shown here.

We see that, for all three centers, the median log-likelihood consistently improves from the first step of adding synthetic data from a single source, but its spread may increase initially, as is the case for \emph{Nottingham}. Spread then diminishes quickly as more synthetic data becomes available. After about five steps only small (but consistent) improvements occur.

\subsection{Data sharing helps especially when local data sets are small}
\label{sec:small_data_not_bad}

We now turn to investigating the effect of the size of the locally available data. Learning of many machine learning models becomes less reliable with smaller amounts of data and the data sharing approach requires each party to learn a generative model under privacy constraints, which poses additional limitations to learning. It is natural to ask whether the quality of the synthetic twin data released by the parties deteriorates more quickly than that of the regression model trained only on local data as the number of data points decreases.

\begin{figure}[t!b]
    \centering
    \frameindraft{\includegraphics[width=.7\columnwidth]{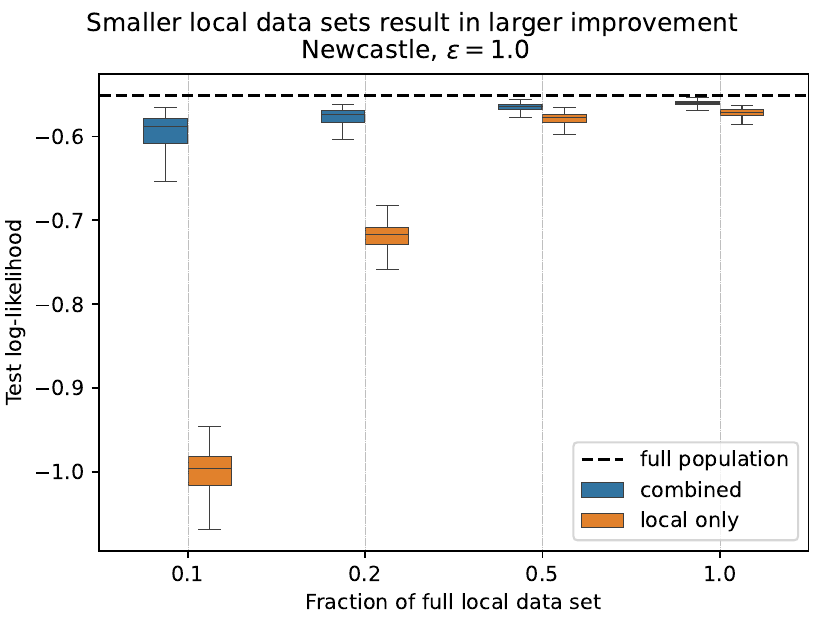}}
    \caption{Usefulness of sharing of data using synthetic twin data sets is retained in the small data regime: Performance of a model trained including shared synthetic twin data from other parties (blue), \emph{all with similarly small local data}, decreases much less than that of a model trained only on locally available data (orange). Higher mean log-likelihoods of combined over local only are statistically highly significant ($p < 0.001, n_{\text{local only}}=\numprint{1000}, n_{\text{combined}}=\numprint{100000}$) for all data set sizes.}
    \label{fig:ll_over_center_sizes}
\end{figure}

To investigate this, in addition to the $10\%$ subsampling used in the earlier experiments we subsample the training data to $20\%$, $50\%$, $100\%$ (i.e., no subsampling) of the original number of samples before running the data sharing procedure, which we again repeat ten times for each setting. Fig.~\ref{fig:ll_over_center_sizes} shows the predictive log-likelihood distribution of training the regression model only on local data (orange) and after including the shared data (blue) for the different amounts of data (at all centers) for the \emph{Newcastle} assessment center, for which the largest amount of local data is available.

We observe that, as the local data gets smaller, the performance of the model trained on local data only deteriorates much faster than that obtained using the data sharing approach. This strongly suggests that the positive effect of getting additional data for the analysis task outpaces the negative effects smaller local data has on learning the generative models. This is most likely due to the negative effects of the latter being mitigated by a sufficient amount of parties sharing data: Even when the individual sets are small and of poor quality, in combination they still carry an overall strong enough signal to enable meaningful analysis. The results for other centers are consistent with this.

\subsection{Parties with large local data can correct for skew in the local distribution}
\label{sec:large_centers_learn}

\begin{figure}[tbh]
    \centering
    \frameindraft{\includegraphics[width=.7\columnwidth]{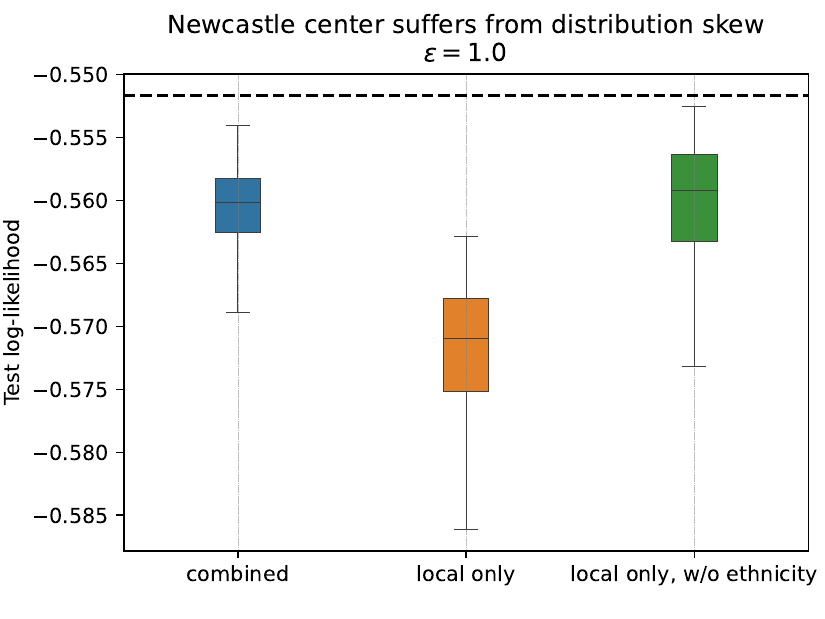}}
    \caption{Training the model on only \emph{Newcastle}'s local data (full data without subsampling) results in poor predictive performance on global data due to skew in the local distribution (orange). Not considering the \emph{ethnicity} feature when training on local data improves predictive performance (green). Combining local data with synthetic twin data also improves model performance while still considering all features, i.e., without need to change the model (blue).
    The dashed line indicates the log-likelihood of a model trained on the full population. Ten independent repeats. Observed pairwise differences between the means of the distributions are statistically highly significant ($p < 0.001, n_{\text{local only}}=\numprint{1000}, n_{\text{combined}}=\numprint{100000}$).}
    \label{fig:loglik_skewed_big_center_gains}
\end{figure}

The final remaining question of interest is whether a large party (i.e., a party with a large amount of local training data) gains anything from engaging in the data sharing procedure. Intuitively, as the local data set of a party $m$ grows, it will reach a point were additional (synthetic) data from other parties will not have a strong effect on the analysis party $m$ performs. Why then should that party participate and share its own data?

To investigate, we isolate the largest assessment center of the UK Biobank data set, \emph{Newcastle}, with \numprint{4737} records in the full training set. Fig.~\ref{fig:loglik_skewed_big_center_gains} shows that contrary to the argument made above, the log-likelihood of the Poisson regression model on the global test set (blue box-plot) is higher than that for the model trained only using Newcastle's local training data (orange).

\subsubsection*{\emph{Newcastle} center is negatively affected by data skew}

\begin{figure}[tb]
    \centering
    \frameindraft{\includegraphics[width=.7\columnwidth]{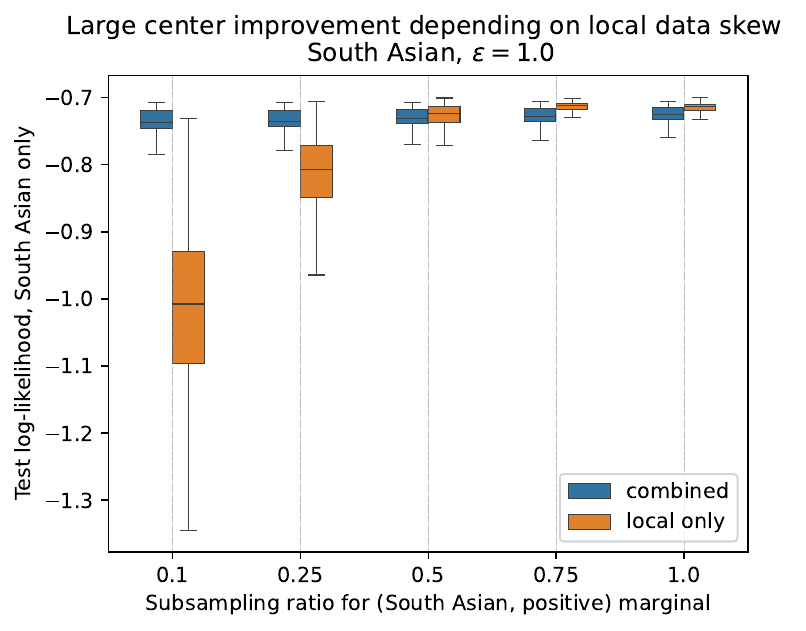}}
    \caption{By participating in the data sharing, parties with large amounts of local data can correct for local skews in data distribution, with stronger corrective effects for stronger skews (corresponding to a smaller value on the x-axis). Observed differences in means are statistically highly significant ($p < 0.001, n_{\text{local only}}=\numprint{1000}, n_{\text{combined}}=\numprint{100000}$) for all skews.}
    \label{fig:loglik_skewed}
\end{figure}

A plausible explanation for this is that the local data is not representative of the global population. It turns out that this particular center is much more ethnically homogeneous, consisting of 96.54 \% records labelled \emph{White British} compared to 88.28 \% in the full UK Biobank cohort. This also manifests in a deviation of the local two-way marginal for the (\emph{ethnicity}, \emph{SARS-CoV-2 test result}) features from the population distribution (cf.~Table~\ref{tab:ukb_ethnicty_covid_two_way_marginal}, App.~\ref{app:ukb_data}).

To determine whether it is this skew that results in the comparatively poor performance of \emph{Newcastle}, we train the regression model on the center's local data without taking ethnicity into consideration as a predictor. This model achieves better predictive accuracy (higher log-likelihood) on the global population (green box-plot of Fig.~\ref{fig:loglik_skewed_big_center_gains}), indicating that this skew in the ethnic composition of the local data does result in an observable effect on global predictions. To learn a good predictive model, the center should therefore not adjust its regression model for ethnicity when training only on local data. However, this means that the model can no longer capture the statistical effect of ethnicity on the outcome. Incorporating shared data from other parties may be able to alleviate the skew in ethnicity, meaning that a model trained on that data would be able to capture its statistical effect. Fig.~\ref{fig:loglik_skewed_big_center_gains} shows that the predictive accuracy of the model trained using shared synthetic data is similar to that of using local data without taking ethnicity into consideration. In this case using combined data achieves a mean log-likelihood slightly worse than the model trained on local data only without considering ethnicity, but does reduce uncertainty and has the additional advantage that the model does not have to be modified.

\subsubsection*{Data sharing mitigates local distribution skew}
To further assess the effect of different magnitudes of local skew, we introduce an artificial new party using half the data from the previously held-out test set. This party's data therefore exactly matches the population distribution and consists of more data points (\numprint{5828}) than any of the original parties. We then fix a category of the \emph{ethnicity} feature, \emph{South Asian}, and discard data points corresponding to the (\emph{South Asian}, \emph{positive SARS-CoV-2 test result}) marginal until only a fraction of $10\%, 25\%, 50\%, 75\%$ remains, respectively, to introduce different amounts of skew.\footnote{In mathematical terms, let $D_{\text{South Asian, positive}} \subset D_m$ denote the subset of data points with \emph{ethnicity} as \emph{South Asian} and a \emph{positive} SARS-CoV-2 test result. $D_{\text{South Asian, negative}} \subset D_m$ is correspondingly the subset of those with \emph{negative} test result. We introduce skew by subsampling $D_{\text{South Asian, positive}}$ with probability $p \in \{0.1, 0.25, 0.5, 0.75, 1\}$ while leaving $D_{\text{South Asian, negative}}$ unchanged.} This experimental set-up corresponds to the effect of minority groups being potentially less likely to report infections, e.g., due to fear of disadvantageous treatment or cases where systematic biases affect the collected data. We evaluate the predictive log-likelihoods on the \emph{South Asian} subgroup of the remaining half of the test set to assess the resulting performance of the regression model for that particular subgroup.

Fig.~\ref{fig:loglik_skewed} shows the predictive log-likelihood distribution for the regression model using only the (skewed) local data (orange) compared to incorporating shared synthetic data (blue) for the different levels of local skew. We observe that the stronger the skew (i.e., the smaller the subsampling ratio of the two-way marginal), the larger the corrective effect from incorporating synthetic data. On the other hand, if only small or no skew is present, the party may not benefit from participating in the sharing. Both of these observations are in line with our expectations.

\section{Discussion}
\label{sec:discussion}

In the above section we have empirically demonstrated on a real-world data set that data sharing using differentially private synthetic twin data from multiple sources improves the performance of the analysis task in many cases. We have particularly shown that the benefits from data sharing diminish more slowly than the deterioration of learning from only local data as the size of the local data set decreases. This is in contrast to the tentative argument made in the beginning that if the local data set is too small to solve the analysis task itself, it will also be too small to learn a good generative model, which would be required for synthetic data sharing to be useful. We believe the reasons for this apparent conflict are twofold: 1) The task faced by the generative model is to learn the \emph{local distribution} of the data, which can be considerably easier, and 2) any remaining errors are sufficiently mitigated by combining synthetic data from multiple sources. This is good news, as it allows data sharing to be adopted in the low data regime, for example, to perform analyses early in an initial data collection phase when only few data points have yet been accumulated at any particular site. The onset period of the SARS-CoV-2 pandemic is an example of exactly such a situation.

We have further seen that the most drastic increase in performance is obtained already with only a small number of parties (5) participating. Additional data from more sources will improve the results further but to a more limited extent, and mostly serves to reduce remaining uncertainty. This is encouraging as it suggests that no large consortium of parties is needed to benefit from the data sharing approach, and any party starting to share their data can do so with the knowledge that only a few like-minded parties are required to reap benefits for all. However, if less than three synthetic twin data sets are made available, there is a risk of low quality synthetic data having a negative impact on the analysis. We discuss this point further in Section~\ref{sec:synthetic_data_can_be_bad}.

Finally, we have experimentally confirmed that parties can successfully correct biases in their data that arise from a local skew of the data distribution, such as the misrepresentation of a minority group. This holds even for the case of parties that already have a large data set, incentivising them to participate and share their data as well.

In the following sections we will discuss variations to the assumptions we made for our experiments (Sec.~\ref{sec:discuss_assumptions}), previous results in the literature about potential harmful effects of using synthetic data (Sec.~\ref{sec:synthetic_data_can_be_bad}), and whether parties could assess whether their analysis improves (Sec.~\ref{sec:discuss_how_do_they_know}). We conclude with a brief discussion of the effect of the privacy parameters (Sec.~\ref{sec:effect_of_epsilon}) and the Federated Learning methodology as a related popular approach for learning from distributed data (Sec.~\ref{sec:federated_learning}).

\subsection{Validity of assumptions}
\label{sec:discuss_assumptions}

In this work we have made several assumptions that we stated in the introduction. We now briefly discuss some possible relaxations and how they would likely affect the results presented above.

We make the simplifying assumption that all parties aim to optimise their analysis on the population level rather than for their local data distribution. If a party instead prefers to optimise for the local distribution there is a high chance that incorporating shared data from other parties decreases this performance (by shifting training the data more towards the overall population distribution) - provided that the party's local distribution deviates from the overall population. On the other hand, for very small local data, the overall population might still add valuable information to aid the learning of the local model. The strength of either effect would likely depend on the amount of deviation and the relative amount of data obtained from other parties. However, the party could easily check whether shared data would improve their analysis by participating in the sharing and simply testing the result on a held-out portion of their data.

We have also assumed that the results of a party's analysis are kept private, i.e., no potential privacy leakage can occur from performing the analysis. If the analysis result is intended to be published, additional measures have to be taken to ensure that the usage of the party's local data does not leak additional data. This could be, e.g., achieved by applying differential privacy in the analysis task or substituting the party's own synthetic twin data in place of its actual data. However, we consider this as an orthogonal problem.

Another assumption we made is that all parties reciprocate in sharing, i.e., if they want to use shared data from other parties, they will also share a synthetic twin of their own data. This seems essential for a fair distribution of the burden of making data available and could be enforced by each party licensing their shared data under conditions that require other parties using it to share alike. However, it is not a strict requirement as long as there is a sufficient number of parties willing to share their data for any reason. Fortunately, as we have seen in the experiments, that number can be quite small and still allow for everyone to see drastic benefits.

Finally, we assumed parties are non-malicious and will not publish data engineered to negatively affect the performance of other parties. Following our procedure a malicious party could sample a large set of arbitrarily bad artificial data to poison the well for all other parties. However, given a sufficient number of non-malicious parties sharing, it is likely that the other parties could filter out such bad shared data sets by comparison with other shared data sets, resulting in some robustness of the overall approach. We consider this to be an interesting future direction.

\subsection{Synthetic data does not unconditionally improve analysis}
\label{sec:synthetic_data_can_be_bad}

Strong privacy protection comes with a trade-off in the usability of the synthetic data. The more privacy we require, the less details we can learn from the data. This trade-off is more severe when data sets are small, because then each individual record has larger effects on the statistics learned from the data, forcing privacy-preserving methods to put stronger restrictions on the learned signals. Hence, statistical signals of the original data tend to become weaker in the synthetic twin data. Additionally, learning a generative model often involves approximations, which can also limit the statistical signal captured in the synthetic twin data. Sampling a finite synthetic data set further introduces uncertainty about the learned parameters of the generative model. 

As a final further complication particular to our setting, the local data contributed by another party could follow a distribution skewed away from that of the overall population - this skew then transfers to the synthetic twin data. Summarising, there are three main factors that can cause synthetic data to be of low quality: 1) the underlying data set was too small to learn a good generative model under differential privacy, 2) the data was skewed away from the population distribution, and, 3) the chosen generative model results in loss of information or skew. Hence, parties interested in the underlying sensitive data might be reluctant to use synthetic twin data for fear of obtaining bad data.

\citet{WildeHJVH21} showed that combining data with a single synthetic data set may lead to worse utility in statistical analysis. Our experiments corroborate their findings and show a large spread in log-likelihood distribution for the analysis task when using data shared by only one or two sources (cf.~Sec.~\ref{sec:more_sharing_better}): In this case there is a relatively high chance of falling victim to a skewed synthetic data set which harms performance of the analysis task compared to only using local data. However, as synthetic data from more and more other centers is combined, the impact of any single bad data set is reduced and we see consistent improvements. This is likely because the data underlying the synthetic twin data becomes a more accurate representation of the population, which eliminates error caused by shift of the local distribution at different centers (Error Source 2). Additional reduction of error may result from averaging out the (largely) independent errors from sources 1) and 3) across multiple synthetic data sets and a general increase in size of available data, although further work is required to confirm this.

We additionally employ a simple technique, drawing inspiration from multiple imputation literature \citep{Rubin04,Reiter07}, to quantify the additional uncertainty introduced by sampling finite (small) data sets from the learned generative models, which is detailed in Sec.~\ref{sec:multiple_imputation}.

\subsection{Can parties evaluate whether shared data improves their model?}
\label{sec:discuss_how_do_they_know}

One of our assumptions in this paper is that all parties aim to optimise the performance of their analysis on the global population. To evaluate this we have used a test set corresponding to the population distribution that we separated from the data prior to running our experiments. However, in practice, parties generally do not have access to an unbiased sample from the population but only their local data. This means it is not trivial for a party to test whether using shared data actually improves their analysis. Testing with a held-out portion of the local data can mislead the party if the local data is skewed away from the population, which the party cannot know a-priori.

As a potential solution to this, parties could establish a joint testing protocol which informs a party of the test performance of their model on population data. This may be based on techniques from the literature on secure multi-party computation \citep{YaoA82,LindellY20} and differential privacy to safeguard learned models and data in the process. Alternatively, an approach using parts of the shared data as a proxy for the population data in testing could be feasible as well. We consider solutions for this an important aspect for future work; here we did not take a stance on what of the alternative solutions is used, and report results assuming it had been solved.

\subsection{Effect of the privacy parameter}
\label{sec:effect_of_epsilon}

The amount of privacy protection afforded by differential privacy is controlled by a privacy parameter $\varepsilon$ (cf.~Sec.~\ref{sec:dp_data_sharing}). All the results shown above were achieved for a fixed amount of $\varepsilon=1$ to achieve comparable results across the different experiments. A stronger level of privacy ($\varepsilon < 1$) would result in a reduction of the information captured in the synthetic twin data. While this decreases analysis performance achieved by the data sharing approach compared to the impractical full pooling of data, a gradual change in $\varepsilon$ does not fundamentally change the overall outcomes and trends we have observed in this paper. We have confirmed this be repeating the experiments with $\varepsilon = 2$ and obtaining similar results with slightly better performance (cf.~Fig.~\ref{fig:opening_result_eps2} in App.~\ref{app:figures}).

\subsection{Federated Learning}
\label{sec:federated_learning}

Federated learning (FL) \citep{McMahan+17} is another collaborative learning framework, where the aim is to learn from distributed data without explicitly combining the data of the parties (typically named \emph{clients} in FL). The clients collaboratively learn a model by locally computing model updates using their own data and sending the updates to an aggregation server which then updates the model's parameters and shares the updated model with the clients. While standard FL does not provide any formal privacy guarantees, there are works merging FL with DP \citep{McMahanRTZ17,Wei+20,Bietti+22}. Although FL (or its DP variants) can provide good performance for certain tasks, it still lacks the generality that synthetic data sharing provides: The data can be used in arbitrary future tasks without any further computational effort or additional expenditure of privacy budget by the parties. FL also requires non-trivial infrastructure: It typically needs a central coordinating party as well as secure real-time two-way communication between that coordinator and all other parties for iterative updates. Our data sharing procedure requires no explicit coordination with other parties, i.e., each party can prepare their synthetic data in a completely \emph{offline} fashion. We only assume that each party has a way of making their synthetic data available to other parties and obtaining synthetic data previously published by the other parties, asynchronously.

However, in cases where combined data is only ever used in a single, well-defined task, an FL approach will likely result in a better privacy--utility trade-off as it can expend all of the available privacy budget on learning that task well. In contrast, our data-sharing-based approach requires each party to learn a model that captures the complete (local) data distribution. FL is also likely preferable for very small local data sets, for which learning a generative model is infeasible.

\section{Methods}
\label{sec:methods}

\subsection{Differentially Private Data Sharing}
\label{sec:dp_data_sharing}

Our approach fundamentally relies on generating synthetic data with privacy guarantees provided by the framework of \emph{differential privacy} (DP) \citep{DworkMNS06}, formally defined as follows:

\begin{definition}[$(\varepsilon, \delta)$ Differential Privacy \citep{DworkMNS06}]
For $\varepsilon \geq 0$ and $\delta \in [0,1]$, a randomised mechanism $\mathcal{M}$ satisfies $(\varepsilon,\delta)$ differential privacy if for any two data sets different in only one element, $\data, \neigh \in \mathcal{D}$, and for all outputs $S \subseteq \text{im}(\mathcal{M})$, the following constraint holds: 
\begin{equation}\label{eq:dp}
    \Pr(\mathcal{M}(\data) \in S) \leq  e^{\varepsilon} \Pr(\mathcal{M}(\neigh) \in S)  + \delta.
\end{equation}
\end{definition}

In this work we have considered the \emph{add-remove} neighbourhood relation, i.e., $\neigh$ can be obtained from $\data$ by adding or removing a single data sample.

Intuitively, the effect of removal or addition of any individual data item in the inputs of a DP algorithm is limited by the privacy parameters $\varepsilon$ and $\delta$. Lower values for these parameters correspond to stricter privacy as they force the output distributions for different inputs to be more similar. In the extreme case of $\varepsilon = 0$ and $\delta = 0$, the output of $\mathcal{M}$ would need to be independent of the inputs.

An important property of differential privacy is \emph{post-processing immunity}, which guarantees that any processing of the outputs of a DP algorithm $\mathcal{M}$ with any function $f$ is still private in the DP sense, i.e., the composition $f \circ \mathcal{M}$ is also $(\varepsilon, \delta)$-DP. Under our assumptions stated in Section~\ref{sec:introduction}, it follows that our framework seen as a whole, which results in each party obtaining a local model trained on the data of all parties, satisfies \emph{billboard differential privacy} \citep{HsuJHRW16, HuSWS23}. This ensures that any party's model is $(\varepsilon, \delta)$-DP with respect to all other parties' data.

\begin{definition}[$(\varepsilon, \delta)$ Billboard Differential Privacy \citep{HuSWS23}]
Let $\mathcal{M}(D) = [f_m(D_m, g(D))]_{m=1,\ldots,M}$ be the output of a randomised mechanism $\mathcal{M}$ with input $D =\bigcup_{m=1}^M D_m \in \mathcal{D}$ where $f_m: \mathcal{D}_m \times \mathcal{Q} \rightarrow \mathcal{R}_m$ and $g: \mathcal{D} \rightarrow \mathcal{Q}$. For $\varepsilon \geq 0$ and $\delta \in [0,1]$, $\mathcal{M}$ satisfies $(\varepsilon, \delta)$ billboard differential privacy if for any $m$ and any two sets $\data_m$ and $\neigh_m$, different in only one element, and for all outputs $S \subseteq \mathcal{R}_{-m} = \bigcup_{j\in \{1,\ldots,M\}\setminus\{m\}} \mathcal{R}_j$, the following constraint holds:
\begin{equation}
    \Pr(\mathcal{M}(\data)_{-m} \in S) \leq e^{\varepsilon} \Pr(\mathcal{M}(\neigh)_{-m} \in S) + \delta,
\end{equation}
where $\mathcal{M}(\cdot)_{-m}$ denotes the output vector of $\mathcal{M}$ with the $m$-th element removed.
\end{definition}

Several methods for the technical implementation of DP data sharing via synthetic data were previously proposed \citep{HardtLM12,ChenAC12,Zhang+14,AcsMCD18,Abay+18,YoonJvdS18,McKennaSM19,Beaulieu+19,HarderAP21,JalkoLHTHK21,RaisaJKH22}. On a high-level, they all specify a $(\varepsilon, \delta)$-DP algorithm $\AlgTrainGen$ for training some generative model $\Gen$ from the sensitive input data and then sample synthetic data from $\Gen$. Due to the post-processing property of DP, the information leakage through the synthetic data is then guaranteed to be bounded by the privacy parameters $\varepsilon$ and $\delta$, regardless of the number of samples drawn from $\Gen$.

In this work we are not suggesting any new inference algorithms for DP data sharing. Instead we suggest a framework for collaborative learning from synthetic data, and we can use any of the aforementioned data generating methods. This allows expert users to choose the most appropriate model for describing their data, which should improve the quality of the shared synthetic data. The particular method and model we used for our experiments are described in the following sections.

\subsection{UK Biobank SARS-CoV-2 data set}
\label{sec:ukb_data_description}

We use a data set obtained from the UK Biobank \citep{Sudlow+15} in our experiments. In particular, we formulate as our analysis task $\Downstream$ a Poisson regression model following \citet{Niedzwiedz+20} which predicts the likelihood of a positive test for a SARS-CoV-2 infection based on five ethnic and socioeconomic factors, among them e.g. an individual's ethnicity and education level.\footnote{The data fields we used from the UK Biobank repository are listed in App.~\ref{app:ukb_data}.} All the features are categorical. The data set we obtained from the UK Biobank is restricted to individuals for which at least one SARS-CoV-2 test result was present (before 2021-03-15) and consists of \numprint{58253} records. These are split over 16 \emph{assessment centers}, in which the individuals signed up for inclusion in the UK Biobank cohort. These splits range in size from \numprint{1867} to \numprint{5922} records, with a median of \numprint{3729}.\footnote{A more detailed listing of center sizes can be found in App.~\ref{app:ukb_data}.} We additionally subsample each split to 10\% of its initial size to create data that is sufficiently small so that a single center cannot learn the regression task well.

We use the assessment centers as parties with their respective split of the full UK Biobank cohort as local data sets $D_m$.

\subsection{Models for analysis task and synthetic data generation}
\label{sec:models}

To formally define the analysis task $\Downstream$, we label the vector of regressors, i.e., the features used for prediction, as $\bx$, the SARS-CoV-2 test result as $y$ and the regression parameters as $\bw$ and obtain $\Downstream$ as:
\begin{align}
   \Downstream(\bx; \bw) &= \arg\max_{y \in \{0,1\}} \frac{\lambda^y e^{-\lambda}}{y!}, \\
   \lambda &= e^{\bw^T \bx}. \label{eq:lambda}
\end{align}
We use the \texttt{statsmodels} \citep{SeaboldSP10} Python package to obtain parameter and corresponding standard error estimates.

For the results presented in Section~\ref{sec:results} we consider a parametric probabilistic model consisting of two parts for generating synthetic data: A mixture model and a Poisson regression. The Poisson regression part exactly mirrors the analysis task and models the SARS-CoV-2 test results based on the regressors. The mixture model part models the regressors following \citet{JalkoLHTHK21}. We use 16 mixture components based on empirical tuning of the model's hyperparameters. Formally, the generative model $\Gen_m$ for each party is:
\begin{align}
    &p(\bx \mid \btheta_{\bx}, \bpi) = 
            \sum_{r=1}^{16} \bpi_r \prod_{j=1}^d p(\bx_j \mid \btheta_j^{(r)}) \\
    &p(y \mid X, \bw) = \frac{\lambda^y \exp(-\lambda)}{y!}.
\end{align}

Here $p(\bx_j \mid \btheta_j^{(r)})$ is the categorical distribution over values of the $j$-th feature of $\bx$ in mixture component $r$ with parameters $\btheta_j^{(r)}$ and $\bpi$ is a vector of mixture coefficients. In our experiments we consider $d=5$ categorical features.

We use the differentially private variational inference algorithm \citep{JalkoDH17} as $\AlgTrainGen$ with $\varepsilon = 1$ and $\delta = \frac{1}{N_m}$ to infer the model parameters for each party in all experiments. $N_m = |D_m|$ is the size of a party's local data set. Synthetic twin data sets (of size $N_m$) are sampled from the inferred model for each party.

\subsection{Sharing multiple synthetic twin data sets}
\label{sec:multiple_imputation}

To quantify additional uncertainty introduced by sampling a finite data set from the generative models we follow an approach suggested by \citet{RaisaJKH22} in which parties perform $K$ repetitions of sampling, publishing and training with synthetic data. Each of the $M$ parties thereby receives $(M-1)K$ synthetic data sets and combines them into $K$ combined sets on which it performs the analysis task. After this, each party locally combines the resulting $K$ predictive models, by either distilling a single combined model out of them or setting up a suitable ensemble. Following \citep{RaisaJKH22}, we use Rubin's rules to combine the obtained parameter and standard error estimates into a single model, analogous to the concept of multiple imputation \citep{Rubin04,Reiter07}, where missing data is repeatedly replaced with resampled available data. We set the number of synthetic data sets sampled by each party as $K=100$. 

\subsection{Evaluation Metrics}
\label{sec:metrics}

We evaluate the regression model $\Downstream$ learned by each party on a test set $D_{\text{all,test}}$ representing the full cohort. $D_{\text{all,test}}$ is obtained by splitting each center's local data into training and test sets with a 80/20 (training/test) ratio, then taking the union over the local test sets to obtain the global test set: $D_{\text{all,test}} = \bigcup_m D_{m,\text{test}}$.\footnote{The local data $D_m$ used by party $m$ for training generative and analysis models then excludes $D_{m,\text{test}}$ in the experiments.}  We use the predictive log-likelihood of $\Downstream$ with learned parameters $\bw$ as our measure for utility:
\begin{align}
    u(\Downstream(\cdot; \bw)) &= \sum_{(\bx, y) \in D_{\text{all,test}}} y\ln\lambda - \lambda - \ln(y!),
\end{align}
with $\lambda$ as defined in Eq.~\eqref{eq:lambda}.

We use a Monte Carlo approach to sample a distribution of log-likelihoods: First we approximate the distribution of parameter estimates by a diagonal Gaussian fully described by the MLE and standard error estimates. We then sample parameter vector $\bw$ from this distribution and compute the corresponding log-likelihood for $\bw$ on $D_{\text{all,test}}$. We repeat this sampling $100$ times. The sampled log-likelihood distribution then summarises the overall performance of the learned regression model across all parameters as well as the estimated standard errors.

We further repeat all experiments 10 times with different seeds for internal randomness, beginning with the inference of the generative models for shared synthetic data. We additionally sample 100 random permutations of the order in which data from other parties becomes available. The plots shown throughout Section \ref{sec:results} always show the distributions of log-likelihood results over all repetitions, for a total of $\numprint{100000}$ samples.

Whenever significance on differences between means of the obtained log-likelihood sample sets were reported, Welch's t-test \citep{WelchB47} was used after ranking the tested samples to account for unequal variances between as well as non-uniformity within each sample set, following \citep{ZimmermanDZ93}. One-sided tests were used in all cases except for Fig.~\ref{fig:loglik_skewed_big_center_gains} and \ref{fig:loglik_skewed}, which used two-sided tests.

\section*{Code Availability}
The code to run the experiments can be found on \url{https://github.com/DPBayes/Collaborative-Learning-with-DP-Synthetic-Twin-Data}.

\section*{Acknowledgement}

This work was supported by the Research Council of Finland (Flagship programme: Finnish Center for Artificial Intelligence, FCAI; and grants 325572, 325573), the Strategic Research Council (SRC) established within the Research Council of Finland (grant 336032), UKRI Turing AI World-Leading Researcher Fellowship (EP/W002973/1), as well as the European Union (Project 101070617). Views and opinions expressed are however those of the author(s) only and do not necessarily reflect those of the European Union or the European Commission. Neither the European Union nor the granting authority can be held responsible for them. This research has been conducted using the UK Biobank Resource under Project Number 65101. This work used data provided by patients and collected by the NHS as part of their care and support (Copyright \textcopyright 2021, NHS England. Re-used with the permission of the NHS England and UK Biobank. All rights reserved). This work used data assets made available by National Safe Haven as part of the Data and Connectivity National Core Study, led by Health Data Research UK in partnership with the Office for National Statistics and funded by UK Research and Innovation (grant MC\_PC\_20058). The authors also acknowledge the computational resources provided by the Aalto Science-IT project.

\printbibliography

\pagebreak
\appendix

\section{UK Biobank Data}
\label{app:ukb_data}

\begin{table}[H]
   \centering
    \caption{Data fields of the UK Biobank used for this work.}
    \begin{threeparttable}
    \begin{tabular}{c|c|c}
         \toprule
         Field ID & Field title & Purpose \\
         \midrule
         31 & Sex & analysis predictor \\
         54 & UK Biobank assessment centre & analysis parties \\
         189 & Townsend deprivation index & analysis predictor \\
         6138 & Qualifications & analysis predictor \\
         21000 & Ethnic background & analysis predictor \\
         21022 & Age at recruitment & analysis predictor \\
         40000 & Date of death & pre-processing (filtering) \\
         40100\tnote{1} & Records of SARS-CoV-2 test results & analysis target \\
         \bottomrule
         \multicolumn{3}{p{251pt}}{%
            Additional filtering and binning of data prior to running the experiments was following the preprocessing steps of \citet{Niedzwiedz+20}.
         }
    \end{tabular}
    \begin{tablenotes}
        \item[1] When data were obtained, SARS-CoV-2 test results were made available as a data table alongside but separate of the main UK Biobank repository, cf.~\url{https://biobank.ctsu.ox.ac.uk/crystal/exinfo.cgi?src=COVID19}.
    \end{tablenotes}
    \end{threeparttable}
   \label{tab:ukb_fields}
\end{table}

\begin{table}[h!]
    \footnotesize
    \centering
    \caption{
        Assessment center data set sizes.
    }
    \begin{tabular}{crr}
        \toprule
        Assessment center & Total data size & Training data size \\
        \midrule
        Newcastle         & \numprint{5922} & \numprint{4737} \\
        Bristol           & \numprint{5860} & \numprint{4688} \\
        Reading           & \numprint{4479} & \numprint{3583} \\
        Leeds             & \numprint{4424} & \numprint{3539} \\
        Bury              & \numprint{4345} & \numprint{3476} \\
        Nottingham        & \numprint{4236} & \numprint{3388} \\
        Hounslow          & \numprint{3984} & \numprint{3187} \\
        Liverpool         & \numprint{3946} & \numprint{3156} \\
        Croydon           & \numprint{3513} & \numprint{2810} \\
        Birmingham        & \numprint{3271} & \numprint{2616} \\
        Sheffield         & \numprint{3042} & \numprint{2433} \\
        Middlesborough    & \numprint{2857} & \numprint{2285} \\
        Stoke             & \numprint{2715} & \numprint{2172} \\
        Barts             & \numprint{1918} & \numprint{1534} \\
        Manchester        & \numprint{1874} & \numprint{1499} \\
        Oxford            & \numprint{1867} & \numprint{1493} \\
        \bottomrule
    \end{tabular}
    \label{tab:ukb_data_center_sizes}
\end{table}

\begin{table}[htb!]
    \footnotesize
    \centering
    \caption{Two-way marginals for ethnicity and SARS-CoV-2 test result, full cohort and \emph{Newcastle} center.
    }
    \begin{tabular}{lrrr}
        \toprule
        Ethnicity & Covid Test Result & Full Cohort & Newcastle \\
        \midrule
        White British & neg. & 19.94 & 18.89 \\
                        & pos.  & 68.33 & 77.64 \\
        \midrule
        White Other     & neg. &  2.38 &  0.92 \\
                        & pos.  &  0.73 &  0.14 \\
        \midrule
        White Irish     & neg. &  2.04 &  0.88 \\
                        & pos.  &  0.61 &  0.19 \\
        \midrule
        South Asian     & neg. &  1.33 &  0.36 \\
                        & pos.  &  0.94 &  0.17 \\
        \midrule
        Black           & neg. &  1.20 &  0.08 \\
                        & pos.  &  0.94 &  0.06 \\
        \midrule
        Other           & neg. &  0.62 &  0.19 \\
                        & pos.  &  0.34 &  0.08 \\
        \midrule
        Mixed           & neg. &  0.41 &  0.27 \\
                        & pos.  &  0.19 &  0.04 \\
        \midrule
        Chinese         & neg. &  0.14 &  0.04 \\
                        & pos.  &  0.05 &  0.0  \\
        \bottomrule
        \multicolumn{4}{p{251pt}}{%
            All values given in per cent. \emph{Newcastle}'s two-way marginals are significantly different to those of the full cohort.
        }
    \end{tabular}
    \label{tab:ukb_ethnicty_covid_two_way_marginal}
\end{table}

\section{Additional Results}
\label{app:figures}

\begin{figure}[H]
    \centering
    \frameindraft{\includegraphics[width=.6\columnwidth]{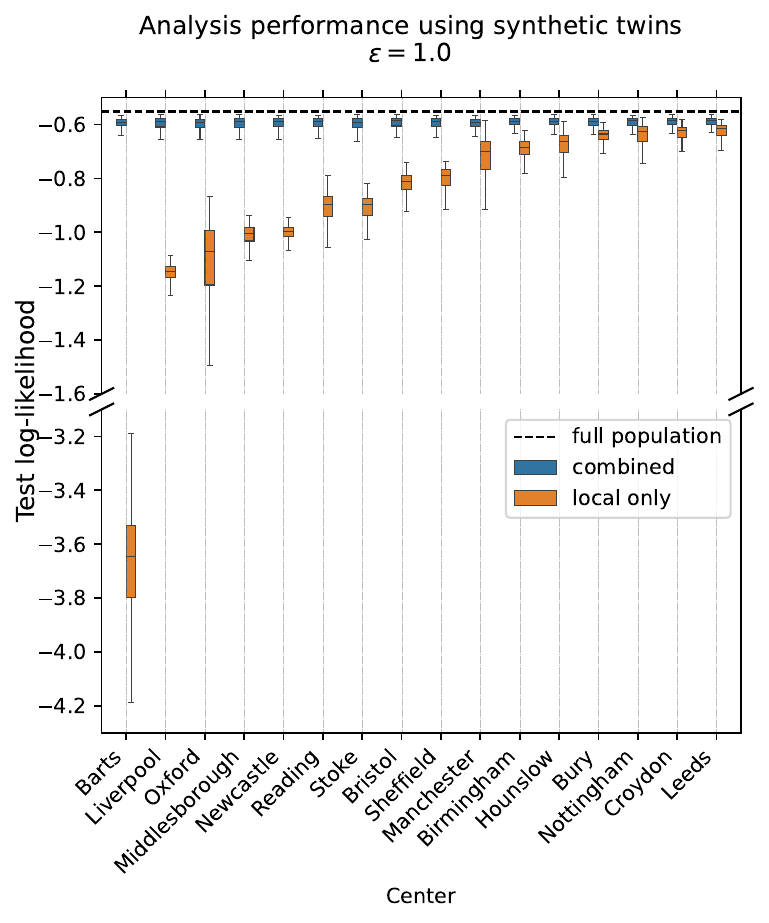}}
    \caption{Full version of right side of Fig.~\ref{fig:abstract_figure_and_opening_result}.}
    \label{fig:full_opening_result}
\end{figure}

\begin{figure}[H]
    \centering
    \frameindraft{\includegraphics[width=.6\columnwidth]{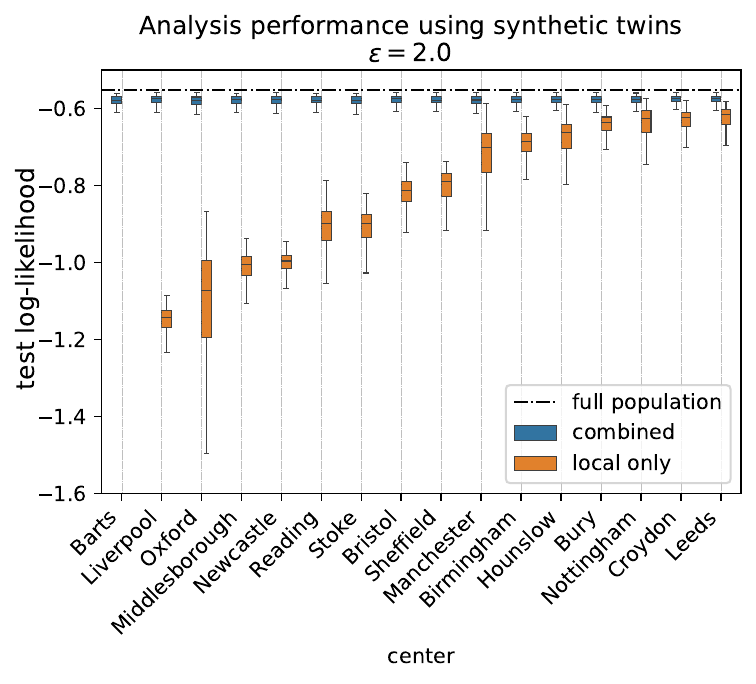}}
    \caption{Right side of Fig.~\ref{fig:abstract_figure_and_opening_result} repeated with privacy parameter $\varepsilon = 2.0$: Higher $\varepsilon$, i.e., less strict privacy guarantees, allows synthetic twin data to carry more information from underlying sensitive data. This reflects in better performance of the model trained with combined data, both in terms of location and spread of the predictive log-likelihood on the test set. The overall trend does not change across privacy levels.}
    \label{fig:opening_result_eps2}
\end{figure}

\begin{table*}[tbh]
\footnotesize
\begin{adjustwidth}{-2cm}{}
    \centering
    \caption{
        p-values for improvements in log-likelihood mean score when adding synthetic twin data from an additional center.
    }
    \begin{tabular}{l|llllllllllll}
        \toprule
        local center & $1 > 0$ & \ldots & $ 6 > 5$ & $7 > 6$ & $8 > 7$ & $9 > 8$ & $10 > 9$ & $11 > 10$ & $12 > 11$ & $13 > 12$ & $14 > 13$ & $15 > 14$ \\
        \midrule
Barts&***& \ldots &***&***&$0.019^{*}$&***&$0.005^{**}$&***&$0.224^{}$&$0.020^{*}$&***&$0.003^{**}$ \\
Birmingham&***& \ldots &***&***&$1.000^{}$&$0.004^{**}$&$0.944^{}$&$0.128^{}$&$0.945^{}$&$0.444^{}$&$0.026^{*}$&$0.285^{}$ \\
Bristol&***& \ldots &***&***&$0.483^{}$&$0.006^{**}$&$0.994^{}$&$0.032^{*}$&$0.990^{}$&$0.419^{}$&$0.021^{*}$&$0.297^{}$ \\
Bury&***& \ldots &***&***&$0.599^{}$&***&$0.674^{}$&$0.039^{*}$&$0.991^{}$&$0.452^{}$&$0.079^{}$&$0.059^{}$ \\
Croydon&***& \ldots &***&$0.004^{**}$&$1.000^{}$&$1.000^{}$&$1.000^{}$&$1.000^{}$&$1.000^{}$&$1.000^{}$&$1.000^{}$&$1.000^{}$ \\
Hounslow&***& \ldots &***&***&$1.000^{}$&$0.012^{*}$&$1.000^{}$&$0.326^{}$&$1.000^{}$&$0.795^{}$&$0.955^{}$&$0.978^{}$ \\
Leeds&***& \ldots &***&***&$1.000^{}$&$0.050^{}$&$1.000^{}$&$0.984^{}$&$1.000^{}$&$0.993^{}$&$0.999^{}$&$1.000^{}$ \\
Liverpool&***& \ldots &***&***&$0.006^{**}$&***&$0.999^{}$&$0.446^{}$&$0.974^{}$&$0.997^{}$&$0.993^{}$&$0.999^{}$ \\
Manchester&***& \ldots &***&***&$0.623^{}$&***&$0.222^{}$&$0.011^{*}$&$0.168^{}$&$0.151^{}$&$0.418^{}$&$0.703^{}$ \\
Middlesborough&***& \ldots &***&***&***&***&$0.006^{**}$&$0.002^{**}$&$0.037^{*}$&$0.212^{}$&$0.291^{}$&$0.330^{}$ \\
Newcastle&***& \ldots &***&***&***&***&***&***&$0.019^{*}$&$0.110^{}$&$0.123^{}$&$0.300^{}$ \\
Nottingham&***& \ldots &***&***&$0.967^{}$&$0.001^{**}$&$0.932^{}$&$0.558^{}$&$1.000^{}$&$0.921^{}$&$0.991^{}$&$0.996^{}$ \\
Oxford&***& \ldots &***&***&***&***&***&***&***&***&***&*** \\
Reading&***& \ldots &***&***&***&***&***&***&***&***&***&$0.002^{**}$ \\
Sheffield&***& \ldots &***&***&$0.356^{}$&$0.006^{**}$&$0.229^{}$&$0.155^{}$&$0.710^{}$&$0.638^{}$&$0.305^{}$&$0.601^{}$ \\
Stoke&***& \ldots &***&***&***&***&***&***&$0.001^{**}$&***&***&$0.001^{**}$ \\
    \bottomrule
    \multicolumn{13}{p{502pt}}{%
         Each row shows the perspective of a local center. Each column reports p-values for the one-sided test that adding another center's data improves the previous mean log-likelihood. Significance levels are $p \leq 0.001$ (\emph{***}), $p \leq 0.01$ (\emph{**}) and $p \leq 0.05$ (\emph{*}). Columns $2 > 1$ to $5 > 4$ with all values $\leq 0.001$ omitted due to space constraints. $n = \numprint{100000}$.
    }
    \end{tabular}
    \label{tab:more_sharing_better_pvalues}
\end{adjustwidth}
\end{table*}

\begin{sidewaysfigure*}
    \centering
    \frameindraft{\includegraphics[width=\textwidth]{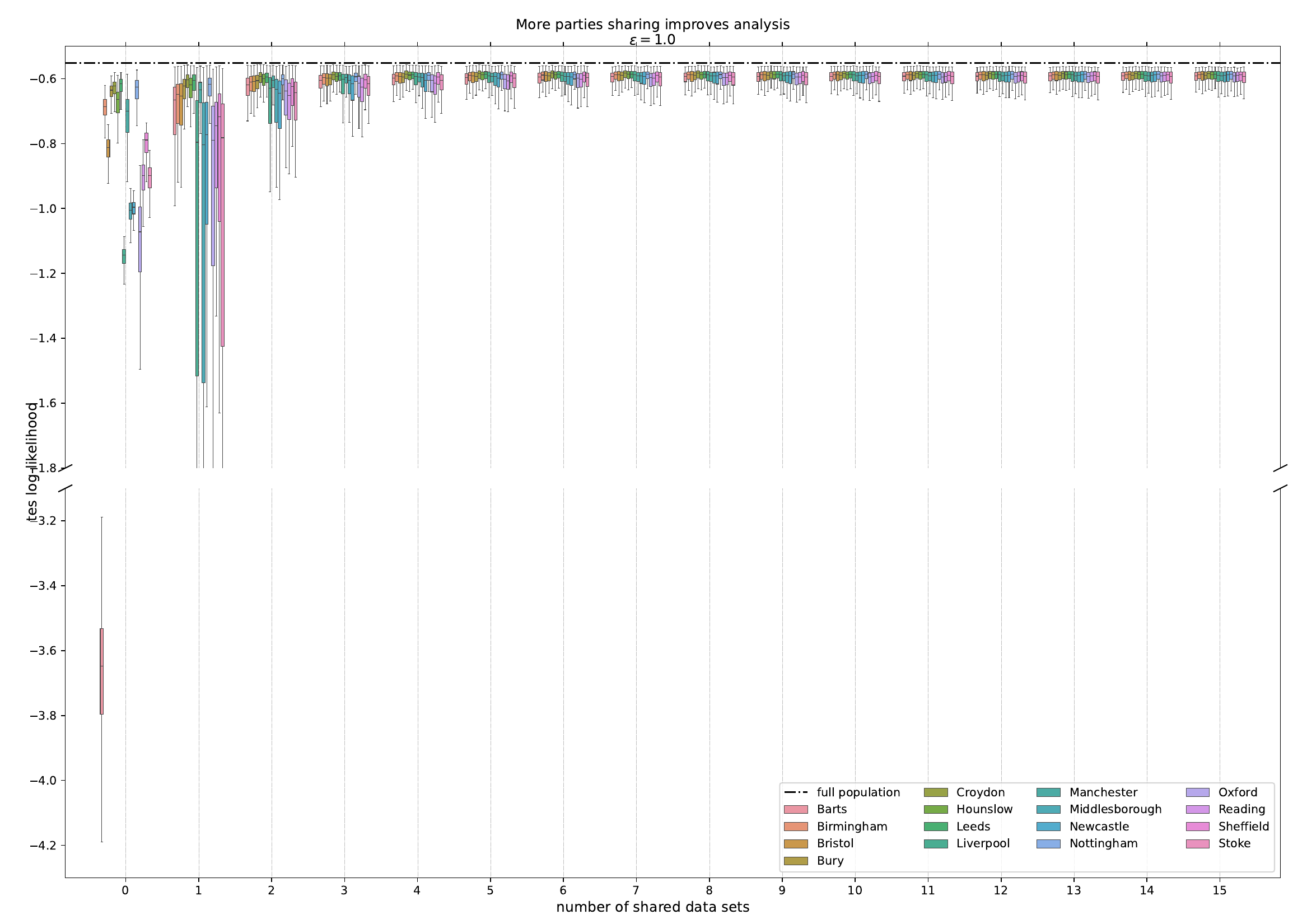}}
    \caption{Full version of Fig.~\ref{fig:ll_over_num_shared}: The log-likelihood of the learned model increases rapidly as synthetic data from other centers becomes available. Spread in log-likelihoods may initially increase when only a few synthetic data sets are incorporated, but then diminishes rapidly with the number of additional data sets. The dashed black line shows the log-likelihood for an ideal setting where the analysis could be performed over the combined data of all parties. 100 repeats with different orders in which synthetic data is added and 10 repeats of releasing synthetic data.}
    \label{fig:full_ll_over_num_shared}
\end{sidewaysfigure*}

\end{document}

%% file: notation.tex
\newcommand{\bx}{\boldsymbol x}
\newcommand{\bw}{\boldsymbol w}

\newcommand{\btheta}{\boldsymbol \theta}

\newcommand{\bpi}{\boldsymbol \pi}

\newcommand{\data}{D} 
\newcommand{\neigh}{D'} 

\newcommand{\Gen}{{G}}
\newcommand{\Downstream}{{f}}
\newcommand{\AlgTrainGen}{\mathcal{T}_\Gen}